\documentclass{article}
\usepackage[utf8]{inputenc}
\usepackage[T1]{fontenc}

\usepackage[margin=1in]{geometry}
\usepackage{authblk}
\usepackage{url}
\usepackage{adjustbox}
\usepackage{caption}
\usepackage{color}
\usepackage{xcolor}
\usepackage[normalem]{ulem}
\usepackage{graphicx}
\usepackage{verbatim}

\usepackage[round,sort&compress,comma,numbers]{natbib}

\usepackage{hyperref}
\usepackage{nameref}
\usepackage{multirow}
\usepackage{booktabs}
\usepackage{indentfirst}
\usepackage[misc]{ifsym}

\usepackage{makecell}

\makeatletter
\renewcommand{\maketitle}{\bgroup\setlength{\parindent}{0pt}
\centering\textbf{\@title}
\begin{flushleft}
  \textbf{Authors:} \@author
\end{flushleft}\egroup
}
\makeatother

\title{\large\textbf{Title: Meta-repository of screening mammography classifiers}}
\author[1*]{Benjamin~Stadnick}
\author[2,3*]{Jan~Witowski}
\author[1*]{Vishwaesh~Rajiv}
\author[4]{Jakub~Chłędowski}
\author[5]{Farah~E.~Shamout}
\author[1,6]{Kyunghyun~Cho}
\author[2,3,7,1,\Letter]{Krzysztof~J.~Geras}

\affil[ ]{\textbf{Affiliations:}}
\affil[1]{Center for Data Science, New York University, NY, USA}
\affil[2]{Department of Radiology, NYU Langone Health, NY, USA}
\affil[3]{Center for Advanced Imaging Innovation and Research, NYU Langone Health, NY, USA}
\affil[4]{Faculty of Mathematics and Information Technologies, Jagiellonian University,~Kraków,~Poland}
\affil[5]{Engineering Division, NYU Abu Dhabi,~Abu Dhabi,~UAE}
\affil[6]{Department of Computer Science, Courant Institute, New York University, NY, USA}
\affil[7]{Vilcek Institute of Graduate Biomedical Sciences, NYU Grossman School of Medicine,~NY,~USA}
\affil[ \Letter ]{\texttt{k.j.geras@nyu.edu}}
\affil[*]{Equal contribution}
\date{}
\begin{document}
\maketitle
\thispagestyle{empty} 
\pagenumbering{arabic}

\noindent\textbf{One Sentence Summary:} We describe a meta-repository of AI models for screening mammography and compare five models on seven international data sets.

\noindent\textbf{Abstract:} Artificial intelligence (AI) is showing promise in improving clinical diagnosis. In breast cancer screening, recent studies show that AI has the potential to improve early cancer diagnosis and reduce unnecessary workup. As the number of proposed models and their complexity grows, it is becoming increasingly difficult to re-implement them. To enable reproducibility of research and to enable comparison between different methods, we release a \textit{meta-repository} containing models for classification of screening mammograms. This meta-repository creates a framework that enables the evaluation of AI models on any screening mammography data set. At its inception, our meta-repository contains five state-of-the-art models with open-source implementations and cross-platform compatibility. We compare their performance on seven international data sets. Our framework has a flexible design that can be generalized to other medical image analysis tasks. The meta-repository is available at \url{https://www.github.com/nyukat/mammography_metarepository}.


\section*{\normalsize INTRODUCTION}
\vspace{-1em}
Mammography is the foundation of breast cancer screening programs around the world~\cite{acr_appropriateness}. Its goal is to reduce breast cancer mortality through early cancer detection. However, it is also known to produce a considerable number of false positive results that lead to unnecessary re-screening and biopsies~\cite{hubbard2011cumulative,gotzsche2013screening}. In addition, mammography can yield false negative diagnosis, resulting in missed early breast cancer detection~\cite{lehman2017national}, especially in patients with high breast tissue density~\cite{nelson2016factors,carney2003individual}.

In recent years, AI has made exceptional progress in cancer detection~\cite{ribli2018detecting,mckinney2020international,jiang2021mri,shen2021artificial,shen2019deep,roela2021deep}. Most recently, the DREAM initiative ran a crowd-sourced challenge~\cite{schaffter2020evaluation} for screening mammography, which included data sets from the United States and Sweden. Over 120 teams from 44 countries participated in the challenge, and the results showed that AI models are on the brink of having radiologist-level accuracy. Although the top-performing models did not beat the radiologists' performance, an ensemble of top-8 models combined with radiologist predictions achieved a superior performance with area under the receiver operating characteristic curve of 0.942 and specificity of 92.0\%. 

A growing interest in this field has resulted in an increase in the number of proposed models and techniques. To realistically quantify the progress and to select the most appropriate solution for a specific application, researchers and clinicians must be able to make fair comparisons between model architectures~\cite{kelly2019key}. AI models are often trained on data sets from populations that differ from the intended target patient group~\cite{rajpurkar2020chexpedition}. At present, it is largely unknown how well proposed models generalize across dissimilar test populations, therefore poor out-of-distribution generalization has been a major concern within the medical AI community~\cite{oakden2020hidden}. The exact definition of the target label might also differ between data sets in subtle, difficult to detect ways. 

Even when there is no explicitly known distribution shift between training and test sets, researchers still face the problem of poor reproducibility. Recent work~\cite{raff2019step} that attempted to reproduce the results of 255 studies was only able to do so in 162 (63.5\%) of them successfully. This is mostly because studies that describe new methods often lack technical details necessary to re-implement them and they do not share their code publicly. That makes it unfeasible to reliably evaluate the performance of those models, and subsequently evaluate how well they generalize to new target populations~\cite{johnson2018reproducibility}. This problem also extends to breast cancer, for which a recent, widely reported study was deemed to be unreproducible~\cite{haibe-kains2020reproducibility}.

For mainstream computer vision tasks, it is straightforward to compare different models using standard, openly available data sets, such as ImageNet~\cite{deng2009imagenet} for image classification and COCO~\cite{lin2014microsoft} for object detection. For less common tasks, such as breast cancer detection on screening mammograms, comparisons between different models are more difficult. The most popular public data sets for screening mammography, DDSM~\cite{heath2000digital} and INbreast~\cite{moreira2012inbreast}, are outdated and small. Equally important, medical data sets have inherent biases based on the acquisition protocols, devices, and the population from which the data was collected. Because of those biases, even if new data sets get published, a definitive analysis of performance for the target clinical setting and all subgroups of interest will remain difficult. As we acknowledge that data sharing is often unfeasible due to legal and ethical restrictions~\cite{krupinski2020ethics,larson2020ethics}, we suggest \textit{model sharing} as a possible substitute. Reproducible models brought together in a single \textit{meta-repository} allow parties of interest to assess standalone model performance. 

\begin{figure}[tb!]
    \includegraphics[width=\textwidth]{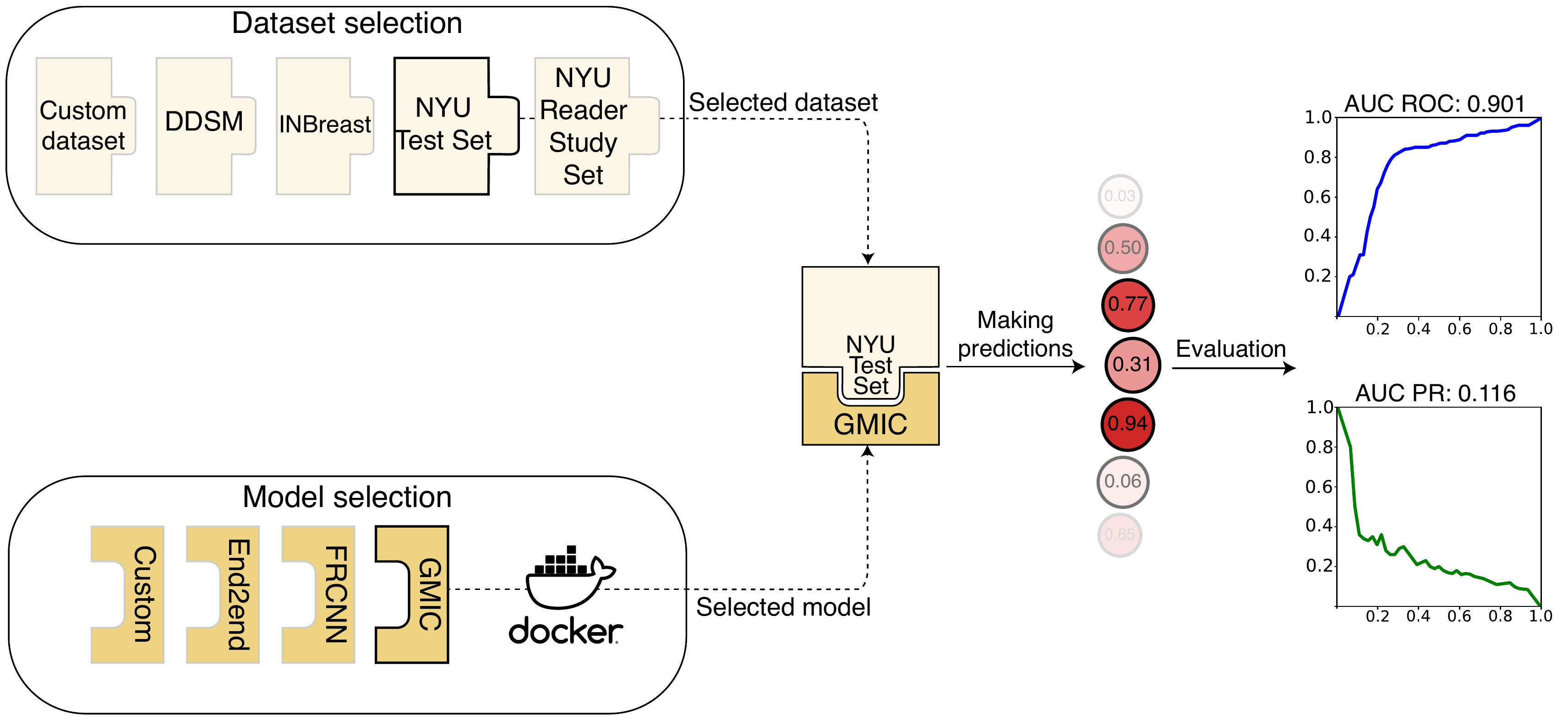}
    \caption{\textbf{An overview of the meta-repository components and functionality.} Containerization of the models enables the reproduction of the results of their official implementations. A selected model is used to generate predictions on a chosen data set, and the meta-repository evaluates the model's performance on this data set. It automatically calculates areas under the receiver operating characteristic curve (AUC ROC) and precision-recall curve (AUC PR) and plots the curves. Models can be evaluated on any public or private mammography data set.}
    \label{fig:overview}
\end{figure}

We hereby introduce a meta-repository that enables researchers to easily evaluate their deep learning models on any public or private screening mammography data set. To overcome problems with reproducibility, we standardize the implementation and evaluation of mammography classifiers and include five open-source, ready-to-use models. Researchers who have access to private data sets can use contributed models to compare their performance on their target population. The meta-repository automatically sets up the environment, evaluates those models on a selected data set, and calculates metrics relevant to breast cancer diagnosis. Finally, in this study we use the meta-repository with the five currently available models to evaluate their performance on seven international data sets in order to assess how well they generalize to different test sets.

We hope that the meta-repository can serve as a hub for breast cancer research, and we encourage others to contribute their models to the repository. Additionally, we include a scoreboard\footnote{Available at the meta-repository website: \url{https://www.github.com/nyukat/mammography_metarepository}} to track the performances of models on the popular public data sets and two private NYU data sets to better measure progress in the field.

\section*{\normalsize RESULTS}
\paragraph{Meta-repository components.}
The meta-repository consists of trained screening mammography classifiers and scripts to load and process data sets, generate predictions, and calculate evaluation metrics (Figure~\ref{fig:overview}). 
Currently, there are five open-sourced screening mammography classifiers ready to use and intended for inference/evaluation only. Everyone is welcome to contribute their models to the meta-repository. To simplify model configuration and enable the use of the meta-repository on various hardware setups, we use Docker~\cite{merkel2014docker} (lightweight software package units, similar to virtual machines) as a tool to wrap the included models. Each model has an associated configuration file for Docker that contains a set of instructions to set up the model's runtime environment. This includes all dependencies, pre-trained weights, scripts, third-party repositories, and toolkits, such as CUDA. Additionally, for each model, it is possible to specify optional parameters for data preprocessing. To evaluate one of the available models, users only have to provide a path to images and a serialized file with ground truth labels. Please refer to Materials and Methods for details on the expected file format.

\begin{figure}[tb!]
    \includegraphics[width=\textwidth]{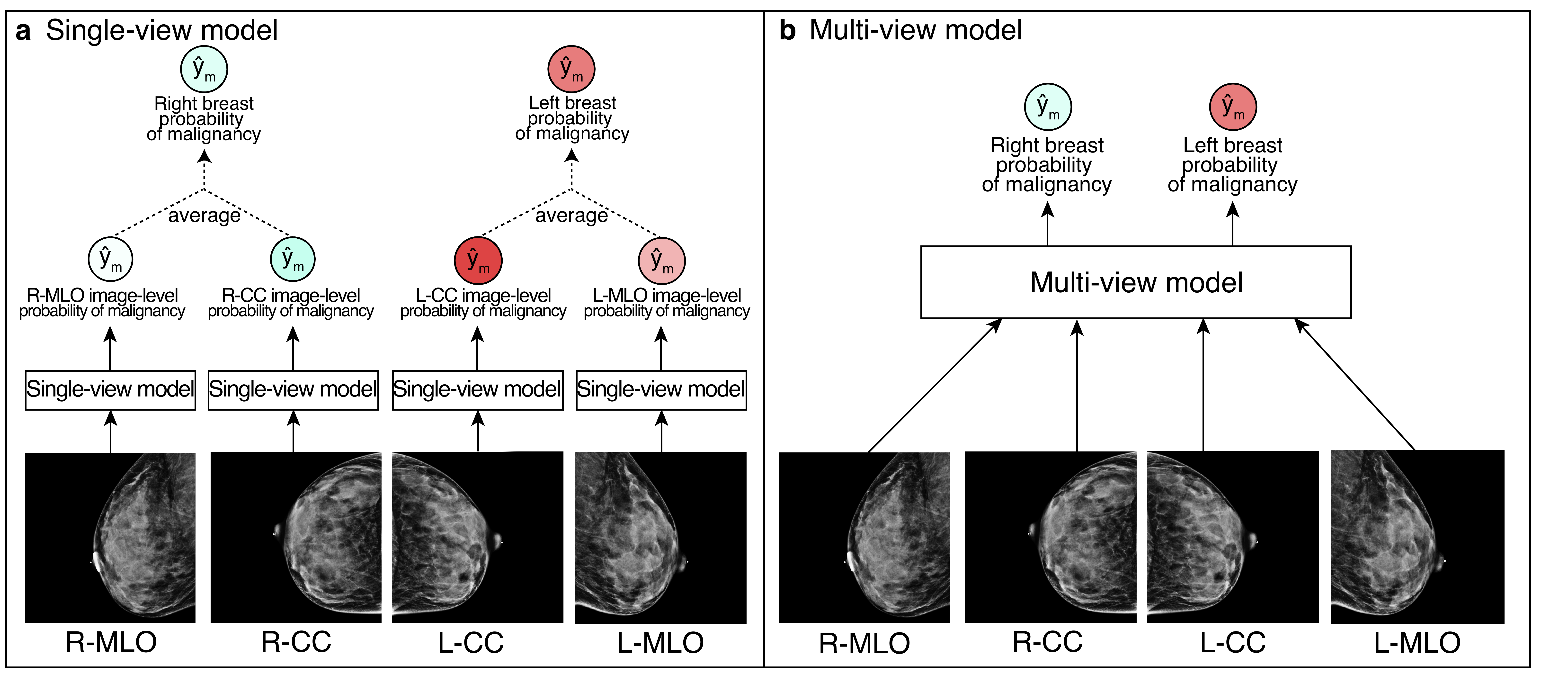}
    \caption{\textbf{Generating breast-level predictions.} General framework for generating breast-level predictions for a single screening mammography exam for \textbf{(a)} single-view and \textbf{(b)} multi-view models. \textit{End2end}, \textit{Faster R-CNN}, \textit{GMIC} and \textit{GLAM} can be considered single-view models, while \textit{DMV-CNN} is by default a multi-view model with an option to work as single-view.}
    \label{fig:multiview_predictions}
\end{figure}

\paragraph{Definition of the prediction task.}
In the meta-repository, we provide models that perform breast cancer screening classification, which evaluates whether or not a malignant lesion is present in the screening mammogram. A model is presented with one or multiple images from an exam. A single image could be one of four standard views that comprise a routine screening mammography exam: left craniocaudal (L-CC), left mediolateral oblique (L-MLO), right craniocaudal (R-CC), right mediolateral oblique (R-MLO). If a model expects a full set of screening mammography images, then it requires one image from each view for both breasts. The model then makes a prediction from the image or set of images (Figure \ref{fig:multiview_predictions}). If one image is given, the model will return an image-level probability of presence of a malignant lesion. If a full set of mammography images is given, the model will instead output two breast-level probabilities: one for the probability of malignancy in the left breast and one for the probability of malignancy in the right breast.

\paragraph{Metrics.}
The area under the receiver operating characteristic curve (AUC ROC) and the area under the precision-recall curve (AUC PR) are standard metrics used to evaluate binary classification tasks and are commonly used in radiology. Precision-recall curves might be more informative than ROC curves in evaluating performance on data sets with a low ratio of positive mammograms (i.e. with malignant findings), and models that optimize AUC ROC do not necessarily optimize the AUC PR~\cite{davis2006relationship}. Although this meta-repository targets populations with the same study indication (screening population), we recognize that many data sets can be and are enriched with more challenging cases or higher disease prevalence than the target population~\cite{gallas2012evaluating}. An example of such a data set is the DDSM test set, which only contains studies with actionable malignant and benign findings. CSAW-CC is also an enriched data set with a limited number of healthy patients. The meta-repository automatically computes AUC ROC and AUC PR at the breast-level and image-level (if applicable). We use a non-parametric approach to estimate AUCs, as it does not make assumptions about the underlying data distribution and therefore has become a standard method in evaluating computer-assisted diagnostic models~\cite{gur2008comparing,kumar2011receiver}. If the model makes predictions for each image (view) separately, then the predictions for each view are averaged together to get the breast-level prediction. Also, plots of the ROC and PR curves are generated and made available to the user. We also calculate and report 95\% bootstrap confidence intervals (N=2,000 bootstrap replicates). If necessary, more metrics can be easily implemented within our evaluation procedure.

\begin{table}[t!]
\caption{\textbf{Breakdown of studies and labels in data sets used to evaluate models' performance.} NYU Reader Study Set is a subset of the NYU Test Set and is the only described data set that allows users to compare model performance to a group of radiologists. The number of studies for DDSM and INbreast data sets represent examples only in the test split, which are subsets of full data sets. CMMD does not have a pre-specified test set, so the table includes all studies in this data set. Numbers for OPTIMAM represent a subset of the full data set to which we were granted access. CSAW-CC does not report benign findings, only whether the case was recalled or not.}
\label{tab:datasets_breakdown}
\centering
\small
\begin{tabular}{@{}lcccccccc@{}}
\toprule
    & \makecell{\textbf{NYU}\\\textbf{Test Set}} & \makecell{\textbf{NYU}\\\textbf{Reader}\\\textbf{Study}}  & \textbf{OPTIMAM} &  & \textbf{DDSM} & \textbf{INbreast} & \textbf{CMMD} & \makecell{\textbf{CSAW-}\\\textbf{CC}} \\ \midrule
    total studies & 14,148 & 720 & 11,633 & & 188 & 31 & 1,775 & 23,395 \\
    \midrule
    malignant           & 40        & 40    & 1,023 & & 89      & 4     & 1,280 &\multirow{2}{*}{524} \\
        benign \& malignant   & 21        & 20    & 175 & & 3       & 11  & 30 &  \\
    benign              & 307       & 300   & 3,829 & & 96      & 16    & 465 & \multirow{2}{*}{22,871}\\
    non-biopsied        & 13,780    & 360   & 6,606 & & 0     & 0   & 0 & \\ \bottomrule
\end{tabular}
\end{table}

\paragraph{Models.}
We identified five models that are ready to use in the meta-repository. These are: \textit{End2end} (Shen et al.~\cite{shen2019deep}\footnote{\url{https://github.com/lishen/end2end-all-conv}}), \textit{Faster R-CNN} (Ribli et al.~\cite{ribli2018detecting}\footnote{\url{https://github.com/riblidezso/frcnn_cad}}), \textit{DMV-CNN} (Wu et al.~\cite{wu2019breastcancer}\footnote{\url{https://github.com/nyukat/breast_cancer_classifier}}),  \textit{GMIC} (Shen et al.~\cite{shen2021interpretable}\footnote{\url{https://github.com/nyukat/GMIC}}), and \textit{GLAM} (Liu et al.~\cite{liu2021weaklysupervised}\footnote{\url{https://github.com/nyukat/GLAM}}). These models were included because they have official, open-source implementations and can be fully evaluated. We will add more models as they become available.

The models differ significantly in terms of architectures and training procedures. For example, \textit{End2end} architecture performs classification on small image patches and then extends the patch classifier to the entire image. \textit{Faster R-CNN} is based on a popular object detection architecture~\cite{girshick2015fast} that identifies suspicious areas in an image and sends them to the classification branch of the network, which identifies them as either benign, malignant, or negative. \textit{DMV-CNN (deep multi-view convolutional neural network)} is the only described model that by default uses all four views simultaneously to make predictions, although a single image version is available in the meta-repository and can be used to evaluate studies with missing views. \textit{GMIC (Globally-Aware Multiple Instance Classifier)} first localizes several regions of interest in the whole image with a global network, then processes extracted regions in a local network, and in the end makes a prediction fusing information from both branches. Finally, \textit{GLAM (Global-Local Activation Maps)} has similar architecture to the \textit{GMIC}, but was designed to produce high-resolution segmentation (which we do not evaluate in this meta-repository) and introduces several changes to the global network. The described variety in model architectures can cause models to learn different image features. Direct comparisons between models might help explore how model design decisions impact their performance.

\begin{table*}[t!]
    \caption{\textbf{Performance across all available models and data sets.} The results show breast-level estimates of areas under the precision-recall curve (AUC PR) and receiver operating characteristic curve (AUC ROC). Results are reported with 95\% confidence interval calculated from bootstrapping with 2,000 bootstrap replicates. For the NYU Reader Study set, we additionally compare model performance against radiologists' results from a 2019 Wu et al.~reader study~\cite{wu2019breastcancer}, which are reported as range (mean). Please refer to Materials and Methods for details regarding parameters used to generate presented results.}
    \centering
    \resizebox{1\textwidth}{!}{
    \begin{tabular}{@{}cccccccccc@{}}
        \toprule
        &&\multicolumn{7}{c}{\textbf{Models}}
        \\ \cmidrule(l){3-9}
        
        \multirow{2}{*}[1.7em]{\textbf{Data set}}
        & \multirow{2}{*}[1.7em]{\textbf{AUC}}
        & \makecell{\textit{End2end}\\(\textit{DDSM})} 
        & \makecell{\textit{End2end}\\(\textit{INbreast})} 
        & \makecell{\textit{Faster}\\\textit{R-CNN}}
        & \textit{DMV-CNN} 
        & \makecell{\textit{GMIC}\\(\textit{single})} 
        & \makecell{\textit{GMIC}\\(\textit{top-5 ensemble})} 
        & \textit{GLAM} 
        & \multirow{2}{*}[1.7em]{\textbf{Radiologists}}  \\
        \midrule
        
        \multirow{2}{*}{\begin{tabular}[c]{@{}c@{}}NYU\\ Reader\\Study\end{tabular}}
        & ROC 
        & \makecell{0.454\\(0.377-0.536)} 
        & \makecell{0.586\\(0.506-0.669)}
        & \makecell{0.714\\(0.643-0.784)} 
        & \makecell{0.779\\(0.719-0.838)} 
        & \makecell{0.857\\(0.802-0.902)} 
        & \makecell{0.881\\(0.836-0.920)}
        & \makecell{0.853\\(0.809-0.894)} 
        & \makecell{0.705-0.860\\(0.778)} \\
        
        & PR  
        & \makecell{0.038\\(0.027-0.051)} 
        & \makecell{0.138\\(0.069-0.229)}
        & \makecell{0.227\\(0.125-0.341)} 
        & \makecell{0.158\\(0.101-0.255)} 
        & \makecell{0.342\\(0.229-0.487)} 
        & \makecell{0.401\\(0.271-0.538)}
        & \makecell{0.229\\(0.137-0.339)} 
        & \makecell{0.244-0.453\\(0.364)} \\
         \midrule
         
        \multirow{2}{*}{\begin{tabular}[c]{@{}c@{}}NYU\\ Test\\Set\end{tabular}}
        & ROC 
        & \makecell{0.475\\(0.403-0.547)} 
        & \makecell{0.608\\(0.523-0.691)}
        & \makecell{0.749\\(0.683-0.810)} 
        & \makecell{0.824\\(0.769-0.877)} 
        & \makecell{0.901\\(0.857-0.939)}
        & \makecell{0.918\\(0.883-0.948)}
        & \makecell{0.887\\(0.848-0.919)} 
        & N/A \\
        
        & PR  
        & \makecell{0.002\\(0.001-0.003)} 
        & \makecell{0.045\\(0.014-0.087)}
        & \makecell{0.033\\(0.011-0.081)} 
        & \makecell{0.027\\(0.012-0.060)} 
        & \makecell{0.116\\(0.052-0.221)}
        & \makecell{0.118\\(0.059-0.212)}
        & \makecell{0.019\\(0.011-0.033)} 
        & N/A \\
        \midrule
        
        \multirow{2}{*}{INBreast}
        & ROC 
        & \makecell{0.676\\(0.469-0.853)} 
        & \makecell{0.977\\(0.931-1.000)}
        & \makecell{0.969\\(0.917-1.000)}
        & \makecell{0.802\\(0.648-0.934)} 
        & \makecell{0.926\\(0.806-1.000)} 
        & \makecell{0.980\\(0.940-1.000)}
        & \makecell{0.612\\(0.425-0.789)} 
        & N/A \\
        
        & PR  
        & \makecell{0.605\\(0.339-0.806)} 
        & \makecell{0.955\\(0.853-1.000)}
        & \makecell{0.936\\(0.814-1.000)}
        & \makecell{0.739\\(0.506-0.906)} 
        & \makecell{0.899\\(0.726-1.000)} 
        & \makecell{0.957\\(0.856-1.000)}
        & \makecell{0.531\\(0.278-0.738)} 
        & N/A \\
        \midrule
        
        \multirow{2}{*}{DDSM}
        & ROC 
        & \makecell{0.904\\(0.864-0.939)}
        & \makecell{0.708\\(0.632-0.782)}
        & N/A* 
        & \makecell{0.532\\(0.456-0.611)} 
        & \makecell{0.604\\(0.522-0.684)} 
        & \makecell{0.610\\(0.529-0.690)}
        & \makecell{0.516\\(0.432-0.596)} 
        & N/A \\
        
        & PR 
        & \makecell{0.891\\(0.836-0.933)}
        & \makecell{0.697\\(0.598-0.789)}
        & N/A* 
        & \makecell{0.461\\(0.373-0.569)} 
        & \makecell{0.541\\(0.436-0.647)} 
        & \makecell{0.569\\(0.470-0.670)}
        & \makecell{0.501\\(0.403-0.603)} 
        & N/A \\
        \midrule
        
        \multirow{2}{*}{CMMD$^\dagger$}
        & ROC 
        & \makecell{0.534\\(0.512-0.557)} 
        & \makecell{0.449\\(0.428-0.473)}
        & \makecell{0.806\\(0.789-0.823)} 
        & \makecell{0.740\\(0.720-0.759)} 
        & \makecell{0.825\\(0.809-0.841)}
        & \makecell{0.831\\(0.815-0.846)}
        & \makecell{0.785\\(0.767-0.803)} 
        & N/A \\
        
        & PR 
        & \makecell{0.517\\(0.491-0.544)} 
        & \makecell{0.462\\(0.438-0.488)}
        & \makecell{0.844\\(0.828-0.859)} 
        & \makecell{0.785\\(0.764-0.806)} 
        & \makecell{0.854\\(0.836-0.869)}
        & \makecell{0.859\\(0.842-0.875)}
        & \makecell{0.818\\(0.798-0.837)} 
        & N/A \\
        \midrule
        
        \multirow{2}{*}{OPTIMAM}
        & ROC 
        & \makecell{0.592\\(0.581-0.602)} 
        & \makecell{0.711\\(0.701-0.720)}
        & \makecell{0.730\\(0.720-0.740)} 
        & \makecell{0.767\\(0.758-0.776)} 
        & \makecell{0.813\\(0.804-0.820)}
        & \makecell{0.832\\(0.825-0.840)}
        & \makecell{0.759\\(0.750-0.768)} 
        & N/A \\
        
        & PR 
        & \makecell{0.266\\(0.254-0.278)} 
        & \makecell{0.437\\(0.421-0.452)}
        & \makecell{0.512\\(0.497-0.526)} 
        & \makecell{0.462\\(0.444-0.477)} 
        & \makecell{0.592\\(0.577-0.608)}
        & \makecell{0.633\\(0.619-0.648)}
        & \makecell{0.488\\(0.470-0.506)} 
        & N/A \\
        
        \midrule
        
        \multirow{2}{*}{CSAW-CC}
        & ROC 
        & \makecell{0.595\\(0.568-0.620)} 
        & \makecell{0.811\\(0.789-0.831)}
        & \makecell{0.909\\(0.896-0.922)} 
        & \makecell{0.872\\(0.855-0.888)} 
        & \makecell{0.933\\(0.921-0.944)}
        & \makecell{0.943\\(0.931-0.954)}
        & \makecell{0.904\\(0.890-0.918)} 
        & N/A \\
        
        & PR 
        & \makecell{0.023\\(0.019-0.031)} 
        & \makecell{0.204\\(0.171-0.240)}
        & \makecell{0.369\\(0.325-0.412)} 
        & \makecell{0.210\\(0.173-0.250)} 
        & \makecell{0.422\\(0.374-0.469)}
        & \makecell{0.495\\(0.447-0.543)}
        & \makecell{0.173\\(0.146-0.206)} 
        & N/A \\
        
        \bottomrule
    \end{tabular}
    }
    \label{tab:model_performance}
    \caption*{\footnotesize$^\ast$Faster R-CNN model was trained on the full DDSM data set, which is why it was not used in this model evaluation. 
    \\$^\dagger$One study ($\#$D1-0951) was excluded for the GMIC model, as its pre-processing failed in this examination.}
\end{table*}

We assessed the models available in the meta-repository on the seven described mammography data sets. The evaluation task was to classify breast-level screening exams as containing malignant lesions or not. Results are reported in section~\nameref{classif_results}.

\paragraph{Data sets.}
To demonstrate the versatility of the meta-repository, we evaluate the aforementioned models on seven data sets (Table \ref{tab:datasets_breakdown}). The first two data sets are from NYU Langone~\cite{wu2019dataset} (the NYU School of Medicine Reader Study Data Set and the NYU School of Medicine Test Data Set). They have been used in previous studies that developed \textit{DMV-CNN}, \textit{GMIC} and \textit{GLAM} models~\cite{wu2019breastcancer, shen2021interpretable, liu2021weaklysupervised, fevry2019improving, makino2020differences,wu2020reducing}. The remaining five data sets are: Digital Database for Screening Mammography (DDSM)~\cite{heath2000digital,lee2017curated}, INbreast~\cite{moreira2012inbreast}, OPTIMAM~\cite{halling2020optimam}, The Chinese Mammography Database (CMMD)~\cite{tcia-cmmd} and CSAW-CC~\cite{dembrower201csaw}. DDSM and INbreast data sets in this study are subsets of the full publicly available data sets. For more information on data sets used in this study, please refer to Materials and Methods.

All cancer cases in described data sets have been pathology-proven, i.e., the diagnosis has been made based on a pathology report, either from a biopsy or surgery. For benign lesions, most cases are also pathology-proven, with the exception of the DDSM data set, where some benign cases were not biopsied. Also, the CSAW-CC data set does not contain information about benign findings, only whether a case was recalled or not.

The data sets come from various institutions around the world. This can be valuable in evaluating model performance on external images in order to prove good model generalizability. Additionally, the meta-repository can be used with private data sets to measure performance in the intended populations. Future public data sets, once available, will also be compatible for use in the meta-repository.

\paragraph{Cancer classification performance.} \label{classif_results}
The results are indicating that models perform better for test sets drawn from the same distribution as the training data. Specifically, on the NYU data sets, the three best performing models were trained on data from the same distribution. \textit{GMIC}, \textit{DMV-CNN} and \textit{GLAM} models achieved performances between 0.779 and 0.857 AUC ROC for the NYU Reader Study set. Meanwhile, \textit{End2end} and \textit{Faster R-CNN} achieved AUC ROC of 0.454 (95\% CI: 0.377-0.536) and 0.714 (0.643-0.784), respectively. The \textit{End2end} model was trained on the DDSM training set, and similarly achieved the best scores on test examples from this data set, with the AUC ROC of 0.904 (0.864-0.939). For the \textit{End2end} model instance fine-tuned on the INbreast dataset, the AUC ROC was 0.977 (0.931-1.000).
It is worth noting that confidence intervals when evaluating on INbreast and DDSM -- the two most popular data sets -- were very wide, which further underlines the necessity of multi-dataset comparisons.

CMMD, OPTIMAM and CSAW-CC are the data sets whose subsets or extended versions have not been used to train any of the models in our metarepository. On CMMD, the best performance was achieved by \textit{GMIC} (0.825 [95\% CI: 0.809-0.841] AUC ROC, 0.854 [0.836-0.869] AUC PR), followed by \textit{Faster R-CNN} (0.806 [95\% CI: 0.789-0.823] AUC ROC, 0.844 [0.828-0.859] AUC PR) and \textit{GLAM} (0.785 [95\% CI: 0.767-0.803] AUC ROC, 0.818 [0.798-0.837] AUC PR). On OPTIMAM, again the best performing model was \textit{GMIC} (0.813 [0.804-0.820] AUC ROC, 0.592 [0.577-0.608] AUC PR), followed by \textit{DMVCNN} (0.767 [0.758-0.776] AUC ROC, 0.462 [0.444-0.477] AUC PR) and \textit{GLAM} (0.759 [0.750-0.768] AUC ROC, 0.488 [0.470-0.506] AUC PR). GMIC was also the strongest model on the CSAW-CC dataset with 0.933 (0.921-0.944) AUC ROC and 0.422 (0.374-0.469) AUC PR.

\textit{GMIC} is the only model in the meta-repository that provides multiple model instances to be used for ensembling. In our experiments, top-5 ensemble outperformed a single \textit{GMIC} instance across all datasets. Also, it reached better scores than the \textit{End2end} model fine-tuned specifically on the INbreast dataset.
Model performance on the NYU Reader Study set can be compared directly to fourteen human experts from a prior reader study~\cite{wu2019breastcancer}. Their AUC ROC ranged between 0.705 and 0.860 with an average of 0.778$\pm$0.04. \textit{GMIC}, \textit{GLAM} and \textit{DMV-CNN} outperform an averaged radiologist AUC ROC. \textit{Faster R-CNN} is slightly trailing the mean AUC ROC, but is still better than the weakest reader. For AUC PR, readers' results varied between 0.244 and 0.453 (average 0.364$\pm$0.05), therefore only the GMIC ensemble of the top-5 models outperformed an average radiologist's AUC PR.

\section*{\normalsize DISCUSSION}

With the presented meta-repository, we aim to facilitate, accelerate, and provide an appropriate benchmark framework for global research on AI in screening mammography. The proposed solution allows researchers to validate their trained models in a fairly fashion. It also gives data owners access to multiple state-of-the-art models. 

To support reproducibility and accessibility of AI in breast cancer imaging research, we included five state-of-the-art open-source models and packaged them with Docker. This ensures that the environment and model implementations are the same for all contributors, enabling fair comparisons between models. Docker containers are lightweight software package units that allow users to follow a specific configuration, install necessary frameworks and run applications on multiple platforms. 

A few projects have attempted to utilize Docker containers in order to facilitate easier access to their models. For example, OncoServe\footnote{\url{https://github.com/yala/OncoServe_Public}} is a repository with three models: one that predicts BI-RADS breast density labels and two detecting probability of developing cancer in the next year or five years given a screening mammogram~\cite{lehman2019mammographic,yala2019deep,yala2019deeptriage}. Another example of using Docker for AI model deployment is ModelHub\footnote{\url{http://modelhub.ai}}, which is a collection of deep learning models for various tasks, including radiological applications~\cite{hosny2019modelhubai}. While our meta-repository also relies on Docker, its purpose is different. Firstly, it is concerned with one task: predicting the probability of breast cancer in a screening mammogram, which means our design is tailored for data and metrics used in this task. Second, our models are all openly available for everyone, unlike OncoServe, which requires an approval from its developers. Third, models used in this meta-repository have been verified by multiple researchers on many datasets. OncoServe developers have only tested their models on a single-vendor data. Finally, both OncoServe and ModelHub have not been updated in over two years. We are committed to maintaining the meta-repository by adding new models, keeping track of the leaderboard, supporting new users and refining the functionalities. 

The presented meta-repository allows us to efficiently compare different models across various public and private data sets. From a comparison perspective, we see that results vary between models. Models that are reported to perform well on the DDSM or INbreast data sets struggle with the NYU data set and vice versa. These discrepancies can be attributed to multiple factors, including the training data used for each model or the exact label definition. These differences highlight the importance and the need for comparison across data sets. To use only one data set for comparison purposes can be misleading. With the meta-repository, we attempt to minimize the effort it takes to run models. Doing this allows for better evaluation/comparison between models as researchers can easily use a model with their own private data set. We hope that other groups will contribute models to the repository. We will maintain a scoreboard for various data sets on the GitHub page of our repository \url{github.com/nyukat/mammography_metarepository}.

\section*{\normalsize MATERIALS AND METHODS} \label{methods}
\subsection*{\normalsize Ethics}
This study, and the processing of the New York University data sets, was approved by the NYU Langone Health Institutional Review Board (approval number s18-00712), and the requirement for informed consent was waived because the study presents no more than minimal risk. All external data sets were fully anonymized before the study was conducted and therefore did not require additional approval. All data processing and experiments were carried out in accordance with the 1964 Helsinki Declaration and its later amendments.

\subsection*{\normalsize Data sets}
\addcontentsline{}{subparagraph}{NYU Test Set}
\addcontentsline{}{subparagraph}{NYU Reader Study Set}

\noindent
\textbf{NYU Test Set} is a subset of the NYU Breast Cancer Screening Dataset~\cite{wu2019dataset}. It was originally used to evaluate the performance of the \textit{DMV-CNN} model. \textbf{NYU Reader Study Set} is a subset of the NYU Test Set, used in the same aforementioned study to compare model performance to a group of radiologists. The Test Set consists of 14,148 breast cancer screening exams of women, ages 27 to 92, and were performed between May and August of 2017 at NYU Langone Health. 307 exams contain only benign lesions, 40 only malignant lesions, and 21 studies both benign and malignant lesions. The remaining 13,780 exams come from patients who did not undergo a biopsy. In the Reader Study Set, there are 720 exams: 300 contain only benign lesions, 40 only malignant lesions, 20 both benign and malignant lesions, and 360 are from patients who did not undergo a biopsy. By including the Reader Study Set as a separate test set, we are able to compare screening mammography model performance directly with radiologists' performance. Fourteen readers (radiologists at various levels of expertise) were asked to provide predictions for all images in this data set.\\

\noindent
The original \textbf{Digital Database for Screening Mammography Test Set (DDSM)} data set was released in 1999 and consisted of 2,620 exams and 10,480 images of digitized film mammograms~\cite{heath2000digital}. Data comes from several institutions in the United States, including Massachusetts General Hospital, Wake Forest University School of Medicine, Sacred Heart Hospital, and Washington University of St. Louis School of Medicine. An updated version of this data set, called the Curated Breast Imaging Subset of DDSM (CBIS-DDSM), was released in 2017 with improved annotations~\cite{lee2017curated}. While CBIS-DDSM specifies a train/test split of the screening mammograms, we used a modified test set that follows Shen et al.~\cite{shen2019deep} to enable comparisons to that study. Please refer to the Supplementary Table~\ref{tab:supplement_ddsm} for a list of patient identifiers used in this test set, which consists of 188 exams. Unlike the NYU data sets, some exams might only have images for selected views or may have images of only one breast. Of the 188 exams, 89 contain only malignant lesions, 96 contain only benign lesions, and the remaining 3 have both malignant and benign lesions. \\

\noindent
The \textbf{INbreast} data set~\cite{moreira2012inbreast} was completed in 2010 and released in 2012. It contains 115 exams from Centro Hospitalar de S. João in Portugal. The data set contains a variety of different lesions with annotations. Additionally, all of the mammograms are digital, as opposed to digitized from film, making this data set the most modern public data set in terms of mammography imaging techniques. The test set we use contains 31 exams from the INbreast data set. Similarly to the DDSM, not all views are present in each exam, and some exams may only consist of one breast. Of the 31 exams, 4 contain only malignant lesions, 16 contain only benign lesions, and 11 have both malignant and benign lesions present. We provide a list of patient identifiers used in the test set in the Supplementary Table~\ref{tab:supplement_inbreast}. \\

\noindent
The \textbf{OPTIMAM}~\cite{halling2020optimam} mammography image database, also known as OMI-DB, includes screening and diagnostic mammograms along with clinical data collected across three UK screening sites since 2011. This data set has restricted access, and only partnering academic groups can gain partial access to the data. In our study, we used a subset of the OPTIMAM database that was shared with us, including 6,000 patients and 11,633 screening mammography studies. OPTIMAM contains images intended for both processing and presentation. In our study, we use only for-presentation images. \\

\noindent
\textbf{The Chinese Mammography Database (CMMD)~\cite{tcia-cmmd}} has been published by The Cancer Imaging Archive (TCIA) in 2021. It includes 1,775 mammography studies from 1,775 patients, acquired between 2012 and 2016, in various Chinese institutions (including Sun Yat-sen University Cancer Center and Nanhai Hospital of Southern Medical University in Foshan~\cite{cai2019breast,wang2016discrimination}). Data set images are accompanied by biopsy-proven breast-level benign and malignant labels. Data set authors also provided age and finding type (calcification, mass or both) for all patients as well as immunohistochemical markers for 749 patients with invasive carcinoma. \\

\noindent
The \textbf{CSAW-CC} data set is a subset of the full CSAW (Cohort of Screen-Aged Women) data set first described in 2019~\cite{dembrower201csaw}. The CSAW-CC is an enriched, case-control (CC) subset based on women who underwent breast screening at the Karolinska University Hospital. It includes all cancer cases and 10,000 randomly selected negative cases. CSAW-CC includes only mammograms acquired on Hologic devices. This data set is dedicated for evaluating AI tools for breast cancer and is available for research purposes from authors. For our analyses, we included only screen-detected cancers and negative cases. \\

\subsection*{\normalsize Usage notes}
Below are usage notes of the meta-repository at the moment of article submission. Please refer to the repository website\footnote{\url{https://github.com/nyukat/mammography_metarepository}} for most up-to-date information.

\paragraph{Images.}
Currently, all of the implemented models in the meta-repository expect 16-bit PNG images as inputs. Mammograms with lower original bit-depth, e.g., 12-bit mammograms, have to be re-scaled to a 16-bit range. Users have to ensure that their images are displayed with presentation intent, meaning that they correctly depict mammograms visually and all necessary grayscale conversions have been applied, such as VOI LUT functions inverting pixel values when using \texttt{MONOCHROME1} Photometric Interpretation. DICOM Standard Section C.8.11.1.1.1 describes in detail differences between for-presentation and for-processing image intent. Image loading is done inside the model's Docker container, so the choice of acceptable image format is left at the developers' discretion. 

\paragraph{Metadata.}
Along with the image files, the meta-repository expects a serialized (``pickled'') metadata file. Metadata files have to be generated for each of the evaluated data sets. Metadata file consists of a list of dictionaries. Each dictionary represents a patient exam and should contain the following information: labels for each breast, image filenames for each of the views, and whether or not images need to be flipped. An example dictionary in the list is shown below:\\
\begin{verbatim}
{
    `L-CC': [`0_L_CC'],
    `R-CC': [`0_R_CC'],
    `L-MLO': [`0_L_MLO'],
    `R-MLO': [`0_R_MLO'],
    `cancer_label': {
        `left_malignant': 0, 
        `right_malignant': 0, 
    }, 
    `horizontal_flip': `NO',
}
\end{verbatim}

There are four keys corresponding to the four standard views in a screening mammography exam: \texttt{L-CC}, \texttt{R-CC}, \texttt{L-MLO}, and \texttt{R-MLO}. Each of these keys has a list of shortened paths as the associated value. These shortened paths can be combined with the image directory path to get the full paths to the images of a specific view for a certain exam. If the exam does not have any images for a view, then the list of shortened paths can be left empty for that key.

The \texttt{cancer\_label} key contains information about exam labels. The value is another dictionary with two keys, \texttt{left\_malignant}, which has a value of 1 if the left breast has a malignant lesion present and a value of 0 if there is no malignant lesion present, and \texttt{right\_malignant}, which has a value of 1 if the right breast has a malignant lesion present and a value of 0 if there is no malignant lesion present. 

The \texttt{horizontal\_flip} attribute is an artifact from the NYU Breast Cancer Screening Dataset~\cite{wu2019dataset} and is currently only used by the NYU models. The NYU models expect the images of each breast to be oriented a certain way. For images of the right breast, the breast should be pointing to the left, i.e.,the sternum/chest should be on the right of the image, and the nipple should be to the left. For images of the left breast, the breast should be pointing to the right. If the images in the exam are oriented this way, then \texttt{horizontal\_flip} is set to \texttt{NO}. If the images in the exam are oriented in the exact opposite manner (images of the right breast point right and images of the left breast point left), then \texttt{horizontal\_flip} is set to \texttt{YES}.

\paragraph{Output.}
Each model saves the predictions into a CSV file whose path is specified at runtime. If the model generates predictions at the image-level, the output should include the following columns: \texttt{image\_index}, \texttt{malignant\_pred}, and \texttt{malignant\_label}. For example:
\begin{verbatim}
    image_index,malignant_pred,malignant_label
    0_L-CC,0.0081,1
    0_R-CC,0.3259,0
    0_L-MLO,0.0335,1
    0_R-MLO,0.1812,0
\end{verbatim}

The \texttt{image\_index} contains the shortened path of the image. The \texttt{malignant\_pred} and \texttt{malignant\_label} columns store the model's prediction for the image and the actual label for the image, respectively. If the model generates predictions at the breast-level, then for each exam, the model needs to generate a left malignant prediction (the probability that a malignant lesion is present in the left breast) and a right malignant prediction (the probability that a malignant lesion is present in the right breast). These predictions are saved to a CSV file with two columns, \texttt{left\_malignant} and \texttt{right\_malignant}, for example:
\begin{verbatim}
    index,left_malignant,right_malignant
    0,0.0091,0.0179
    1,0.0012,0.7258
    2,0.2325,0.1061
\end{verbatim}

The order of predictions is assumed to match the order of the exams in the pickle file, and labels are extracted from the pickle file when scoring occurs.

\paragraph{Model optional parameters.}
\noindent
To replicate the performance as described in the Table~\ref{tab:model_performance}, please keep in mind the following usage notes for specific models:
\begin{itemize}
    \item For the \textit{End2end} model, there is a number of pretrained models available. In our study, we report results on two \textit{End2end} model instances: ResNet50-based model trained on the DDSM data, as well as the VGG-16-based model trained on DDSM and fine-tuned on the INbreast data set. In our meta-repository, there are more \textit{End2end} pretrained models available with various architectures, which in our experiments yielded lower performance.
    \item For the \textit{End2end} model, one should update the mean pixel intensity parameter in the configuration file. By default, this parameter has a value of 44.4, which is a mean pixel intensity value for the INbreast data set. For evaluation on the other data sets, this value should be set to: DDSM - 52.18, CMMD - 18.01, NYU (for Reader Study and Test Set) - 31.28, OPTIMAM - 35.15, CSAW-CC - 23.14. For other data sets, one needs to find the mean pixel intensity value and set it appropriately.
    \item There are no optional configuration parameters for the \textit{Faster R-CNN} model.
    \item The \textit{DMV-CNN} model by default expects all four views to be present in each exam. If the data set does not have all 4 views present for each exam, one must use a special instance of the \textit{DMV-CNN} model, which is defined by a different argument (\texttt{nyu\_model\_single} instead of \texttt{nyu\_model}). This is the case for all data sets described in this study.
    \item There are weights of five pre-trained classifiers available for the \textit{GMIC} model. All of those models have the same architecture, but they have been trained with various hyperparameters during random search, and authors provide top 5 models. By default, classifier number 1 (top 1) is selected. There is also an option to select an ensemble of all top 5 models. Performance in the Table~\ref{tab:model_performance} is reported for the top 1 model only.
    \item The \textit{GLAM} model has two versions of this model with separate pre-trained weights: \texttt{model\_sep} and \texttt{model\_joint}. By default, meta-repository uses \texttt{model\_joint} weights. Please refer to the original paper~\cite{liu2021weaklysupervised} to learn more details about differences between joint and separate training of \textit{GLAM}. In Table~\ref{tab:model_performance}, we report performance for the \texttt{model\_joint} across all data sets.
\end{itemize}

\subsection*{\normalsize Contributing new models}
Authors who wish to include their models in the meta-repository must create a pull request, in which they provide a Dockerfile, an entrypoint bash script, and a sample configuration file. Examples are available online in the metarepository. The Dockerfile should recreate the environment used in the original implementation as closely as possible. The entrypoint will be called at runtime with a set of arguments specifying, among other things, the path to the image directory, the device to use (GPU or CPU), and the path where the output file should be saved. This script should execute all necessary files that generate predictions and/or perform pre or postprocessing. The model should produce an output CSV file following the standard format specified in the README file of our repository for image-level or breast-level predictions depending on the type of predictions the model makes.

\subsection*{\normalsize Submission policy}
We will evaluate each model contributed to the meta-repository up to three times on all data sets. We offer the two extra evaluations to enable some minimal debugging. We will also consider evaluating private models not contributed to the meta-repository on a case-by-case basis.

\subsection*{\normalsize Supplementary Materials}
\begin{itemize}
    \item[] Table S1. Patient identifiers from the DDSM data set.
    \item[] Table S2. Image names of the test examples from the INbreast data set.
\end{itemize}

\bibliographystyle{ieeetr}
\bibliography{mybib.bib}
\appendix

\paragraph{\textbf{Acknowledgements:}} The authors would like to thank Mario Videna, Abdul Khaja and Michael Costantino for supporting our computing environment. The authors would also like to thank Miranda Pursley for proofreading the manuscript. 

\paragraph{\textbf{Funding:}}
\begin{itemize}
    \item[] National Institutes of Health grant P41EB017183 (KJG)
    \item[] National Institutes of Health grant R21CA225175 (KJG, JW, KC)
    \item[] Gordon and Betty Moore Foundation grant 9683 (KJG)
    \item[] Polish National Agency for Academic Exchange grant PPN/IWA/2019/1/00114/U/00001 (JW)
    \item[] Polish National Science Center grant 2021/41/N/ST6/02596 (JC)
\end{itemize}

\paragraph{\textbf{Author contributions:}} 
\begin{itemize}
    \item[] Conceptualization: BS, VR, KC, KG
    \item[] Software development: BS, VR, JC, JW
    \item[] Data analysis: BS, VR, FS, JW, KG
    \item[] Result evaluation: BS, VR, FS, JW, KG
    \item[] Project management: JW, KC, KG
    \item[] Supervision: KC, KG
    \item[] Writing - original draft: BS, JW, JC, FS, KG
    \item[] Writing - review \& editing: BS, JW, VR, JC, FS, KC, KG
\end{itemize}

\paragraph{\textbf{Competing interests:}} Authors declare that they have no competing interests.

\paragraph{\textbf{Data availability:}} Three of the described data sets, DDSM, INbreast and CMMD, contain de-identified images and are publicly available to download. INbreast data set is no longer updated by the original research group, but authors of the data set encourage users to contact them to gain access to images~\footnote{\url{https://github.com/wentaozhu/deep-mil-for-whole-mammogram-classification/issues/12\#issuecomment-1002543152}}. DDSM and its updated version, CBIS-DDSM, are available in the Cancer Imaging Archive (\url{https://doi.org/10.7937/K9/TCIA.2016.7O02S9CY}). CMMD is also available through TCIA (\url{https://doi.org/10.7937/tcia.eqde-4b16}). The OPTIMAM and CSAW-CC data sets are restricted access databases, and interested users should contact OPTIMAM and/or CSAW-CC investigators with an access request. While the two remaining NYU data sets are not publicly available for download, we will evaluate models on the NYU data upon their submission to the metarepository.

\paragraph{\textbf{Code availability:}} We publicly release full code of the meta-repository at \url{https://github.com/nyukat/mammography_metarepository}. The included models are also open-sourced and publicly available: \textit{GMIC} - \url{https://github.com/nyukat/GMIC}; \textit{GLAM} - \url{https://github.com/nyukat/GLAM}; \textit{DMV-CNN} - \url{https://github.com/nyukat/breast_cancer_classifier}; \textit{Faster R-CNN} - \url{https://github.com/riblidezso/frcnn_cad}; \textit{End2end} - \url{https://github.com/lishen/end2end-all-conv}. In our meta-repository, we made use of several open-source libraries and frameworks, such as PyTorch (\url{https://pytorch.org}) machine learning framework.

\clearpage
\section*{Supplementary information} \label{supplement_section}
\renewcommand{\tablename}{Supplementary Table}
\setcounter{table}{0}
\begin{table}[ht]
\begin{center}
    
\footnotesize
    \begin{tabular}{|l|l|l|l|l|l|l|l|}
         \hline 
         P\_02409 & P\_01240 & P\_00726 & P\_01730 & P\_01326 & P\_01633 & P\_00473 & P\_00607 \\
         P\_01442 & P\_01229 & P\_00931 & P\_01379 & P\_01890 & P\_00529 & P\_01721 & P\_01457 \\
         P\_01823 & P\_01420 & P\_00078 & P\_01522 & P\_01585 & P\_00373 & P\_00880 & P\_00841 \\
         P\_00855 & P\_00491 & P\_01545 & P\_00616 & P\_01338 & P\_00430 & P\_02079 & P\_01293 \\
         P\_01188 & P\_00422 & P\_01383 & P\_00063 & P\_00413 & P\_01732 & P\_00254 & P\_00275 \\
         P\_01689 & P\_00625 & P\_01543 & P\_01636 & P\_00923 & P\_00624 & P\_00276 & P\_01184 \\
         P\_00107 & P\_00101 & P\_00224 & P\_01710 & P\_00496 & P\_01342 & P\_00159 & P\_00658 \\
         P\_00630 & P\_01663 & P\_01040 & P\_00455 & P\_00488 & P\_01179 & P\_01888 & P\_01196 \\
         P\_00058 & P\_00420 & P\_00586 & P\_00349 & P\_01635 & P\_00682 & P\_00480 & P\_01566 \\
         P\_00858 & P\_00666 & P\_01264 & P\_00081 & P\_00967 & P\_00250 & P\_01885 & P\_00969 \\
         P\_01630 & P\_00775 & P\_00438 & P\_00427 & P\_02092 & P\_00890 & P\_01736 & P\_02460 \\
         P\_00534 & P\_02489 & P\_00685 & P\_01252 & P\_00592 & P\_01766 & P\_01777 & P\_01239 \\
         P\_00733 & P\_01343 & P\_00441 & P\_00462 & P\_01057 & P\_00591 & P\_00137 & P\_00019 \\
         P\_00582 & P\_00018 & P\_01258 & P\_01775 & P\_01384 & P\_00787 & P\_00400 & P\_00188 \\
         P\_01642 & P\_00605 & P\_01644 & P\_01575 & P\_01474 & P\_01155 & P\_01314 & P\_00863 \\
         P\_00517 & P\_01760 & P\_00332 & P\_00342 & P\_00023 & P\_00709 & P\_01149 & P\_00355 \\
         P\_00321 & P\_01036 & P\_00136 & P\_00146 & P\_00411 & P\_01385 & P\_01225 & P\_00836 \\
         P\_01669 & P\_00563 & P\_02437 & P\_00864 & P\_01444 & P\_00239 & P\_01577 & P\_01983 \\
         P\_01319 & P\_01718 & P\_00320 & P\_00613 & P\_00452 & P\_00819 & P\_00743 & P\_00581 \\
         P\_01047 & P\_00673 & P\_00263 & P\_00515 & P\_01120 & P\_02259 & P\_01260 & P\_01678 \\
         P\_00015 & P\_00185 & P\_01596 & P\_01856 & P\_01810 & P\_01650 & P\_01753 & P\_02563 \\
         P\_01332 & P\_01052 & P\_01517 & P\_00678 & P\_01203 & P\_01367 & P\_02220 & P\_01569 \\
         P\_01434 & P\_00012 & P\_00039 & P\_00314 & P\_00665 & P\_01560 & P\_00997 & P\_01219 \\
         P\_00225 & P\_01439 & P\_00808 & P\_00705 & & & &  \\ \hline 
    \end{tabular}
    \caption{Patient identifiers from the DDSM data set that consist a test set, as described previously in Shen et al.~\cite{shen2019deep}.}
    \label{tab:supplement_ddsm}
\end{center}
\end{table}

\begin{table}[ht]
\begin{center}
\footnotesize
    \begin{tabular}{|l|l|}
         \hline 
         neg/20588138\_8d0b9620c53c0268\_MG\_R\_ML\_ANON & neg/20588164\_8d0b9620c53c0268\_MG\_R\_CC\_ANON \\neg/20588458\_bf1a6aaadb05e3df\_MG\_R\_CC\_ANON & neg/20588510\_bf1a6aaadb05e3df\_MG\_R\_ML\_ANON \\neg/20588654\_036aff49b8ac84f0\_MG\_R\_ML\_ANON & neg/22580492\_2a5b932da4ce5ca1\_MG\_R\_CC\_ANON \\neg/22580548\_2a5b932da4ce5ca1\_MG\_R\_ML\_ANON & neg/22613796\_45c7f44839fd9e68\_MG\_L\_CC\_ANON \\neg/22613848\_45c7f44839fd9e68\_MG\_L\_ML\_ANON & neg/22613944\_f23fa352e7de3dc7\_MG\_L\_CC\_ANON \\neg/22613996\_f23fa352e7de3dc7\_MG\_L\_ML\_ANON & neg/22614353\_d065adcb9905b973\_MG\_R\_CC\_ANON \\neg/22614405\_d065adcb9905b973\_MG\_R\_ML\_ANON & neg/22670124\_e1f51192f7bf3f5f\_MG\_L\_CC\_ANON \\neg/22670177\_e1f51192f7bf3f5f\_MG\_L\_ML\_ANON & neg/22670301\_98429c0bdf78c0c7\_MG\_L\_CC\_ANON \\neg/22670347\_98429c0bdf78c0c7\_MG\_L\_ML\_ANON & neg/22678980\_b9a4da5f2dae63a9\_MG\_L\_CC\_ANON \\neg/22679036\_b9a4da5f2dae63a9\_MG\_L\_ML\_ANON & neg/24065680\_5291e1aee2bbf5df\_MG\_L\_ML\_ANON \\neg/24065734\_5291e1aee2bbf5df\_MG\_L\_CC\_ANON & neg/30011647\_6968748e66837bc7\_MG\_R\_CC\_ANON \\neg/30011700\_6968748e66837bc7\_MG\_R\_ML\_ANON & neg/30011798\_4f20c1285d8f0b1f\_MG\_R\_CC\_ANON \\neg/30011850\_4f20c1285d8f0b1f\_MG\_R\_ML\_ANON & neg/50993616\_b03f1dd34eb3c55f\_MG\_L\_ML\_ANON \\neg/50993643\_b03f1dd34eb3c55f\_MG\_L\_CC\_ANON & neg/50993670\_b03f1dd34eb3c55f\_MG\_L\_ML\_ANON \\neg/50993697\_b03f1dd34eb3c55f\_MG\_L\_CC\_ANON & neg/50995872\_c94d8a1ebd452afe\_MG\_L\_ML\_ANON \\neg/50995899\_c94d8a1ebd452afe\_MG\_L\_CC\_ANON & neg/50995963\_d742ec2f9b90aa62\_MG\_L\_ML\_ANON \\neg/50995990\_d742ec2f9b90aa62\_MG\_L\_CC\_ANON & neg/50996201\_8c1b2bd64ca4d778\_MG\_L\_ML\_ANON \\neg/50996228\_8c1b2bd64ca4d778\_MG\_L\_CC\_ANON & neg/50996709\_330e5fe16929eed4\_MG\_R\_ML\_ANON \\neg/50996736\_330e5fe16929eed4\_MG\_R\_CC\_ANON & neg/50996800\_fdf4a1516f88b280\_MG\_L\_ML\_ANON \\neg/50996827\_fdf4a1516f88b280\_MG\_R\_ML\_ANON & neg/50996854\_fdf4a1516f88b280\_MG\_L\_CC\_ANON \\neg/50996881\_fdf4a1516f88b280\_MG\_R\_CC\_ANON & neg/50998032\_66adfbb4f19c76d2\_MG\_R\_CC\_ANON \\neg/50998086\_66adfbb4f19c76d2\_MG\_R\_ML\_ANON & neg/50999094\_cb65e8dac169f596\_MG\_L\_ML\_ANON \\neg/50999121\_cb65e8dac169f596\_MG\_R\_ML\_ANON & neg/50999148\_cb65e8dac169f596\_MG\_L\_CC\_ANON \\neg/50999175\_cb65e8dac169f596\_MG\_R\_CC\_ANON & neg/50999246\_cb65e8dac169f596\_MG\_L\_ML\_ANON \\neg/50999273\_cb65e8dac169f596\_MG\_R\_ML\_ANON & neg/50999300\_cb65e8dac169f596\_MG\_L\_CC\_ANON \\neg/50999327\_cb65e8dac169f596\_MG\_R\_CC\_ANON & neg/51048891\_f3e93e889a7746f0\_MG\_L\_ML\_ANON \\neg/51048918\_f3e93e889a7746f0\_MG\_R\_ML\_ANON & neg/51048945\_f3e93e889a7746f0\_MG\_L\_CC\_ANON \\neg/51048972\_f3e93e889a7746f0\_MG\_R\_CC\_ANON & neg/51049249\_832ebce700241036\_MG\_L\_CC\_ANON \\neg/51049276\_832ebce700241036\_MG\_L\_ML\_ANON & neg/53582395\_3f0db31711fc9795\_MG\_L\_ML\_ANON \\neg/53582422\_3f0db31711fc9795\_MG\_R\_ML\_ANON & neg/53582449\_3f0db31711fc9795\_MG\_L\_CC\_ANON \\neg/53582476\_3f0db31711fc9795\_MG\_R\_CC\_ANON & neg/53586361\_dda3c6969a34ff8e\_MG\_L\_ML\_ANON \\neg/53586388\_dda3c6969a34ff8e\_MG\_R\_ML\_ANON & neg/53586415\_dda3c6969a34ff8e\_MG\_L\_CC\_ANON \\neg/53586442\_dda3c6969a34ff8e\_MG\_R\_CC\_ANON & neg/53586724\_e5f3f68b9ce31228\_MG\_L\_ML\_ANON \\neg/53586751\_e5f3f68b9ce31228\_MG\_R\_ML\_ANON & neg/53586778\_e5f3f68b9ce31228\_MG\_L\_CC\_ANON \\neg/53586805\_e5f3f68b9ce31228\_MG\_R\_CC\_ANON & neg/53586869\_6ac23356b912ee9b\_MG\_L\_ML\_ANON \\neg/53586896\_6ac23356b912ee9b\_MG\_L\_CC\_ANON & neg/53587104\_7b71aa9928e6975e\_MG\_L\_ML\_ANON \\neg/53587131\_7b71aa9928e6975e\_MG\_L\_CC\_ANON & neg/53587663\_5fb370d4c1c71974\_MG\_R\_CC\_ANON \\neg/53587690\_5fb370d4c1c71974\_MG\_L\_ML\_ANON & neg/53587717\_5fb370d4c1c71974\_MG\_R\_ML\_ANON \\neg/53587744\_5fb370d4c1c71974\_MG\_L\_CC\_ANON & pos/20587054\_b6a4f750c6df4f90\_MG\_R\_CC\_ANON \\pos/20587080\_b6a4f750c6df4f90\_MG\_R\_ML\_ANON & pos/20588190\_8d0b9620c53c0268\_MG\_L\_CC\_ANON \\pos/20588216\_8d0b9620c53c0268\_MG\_L\_ML\_ANON & pos/20588536\_bf1a6aaadb05e3df\_MG\_L\_ML\_ANON \\pos/20588562\_bf1a6aaadb05e3df\_MG\_L\_CC\_ANON & pos/22580520\_2a5b932da4ce5ca1\_MG\_L\_CC\_ANON \\pos/22580576\_2a5b932da4ce5ca1\_MG\_L\_ML\_ANON & pos/22613770\_45c7f44839fd9e68\_MG\_R\_CC\_ANON \\pos/22613822\_45c7f44839fd9e68\_MG\_R\_ML\_ANON & pos/22613918\_f23fa352e7de3dc7\_MG\_R\_CC\_ANON \\pos/22613970\_f23fa352e7de3dc7\_MG\_R\_ML\_ANON & pos/22614236\_1e5c3af078f74b05\_MG\_L\_CC\_ANON \\pos/22614266\_1e5c3af078f74b05\_MG\_L\_ML\_ANON & pos/22614379\_d065adcb9905b973\_MG\_L\_CC\_ANON \\pos/22614431\_d065adcb9905b973\_MG\_L\_ML\_ANON & pos/22670094\_e1f51192f7bf3f5f\_MG\_R\_CC\_ANON \\pos/22670147\_e1f51192f7bf3f5f\_MG\_R\_ML\_ANON & pos/22670278\_98429c0bdf78c0c7\_MG\_R\_CC\_ANON \\pos/22670324\_98429c0bdf78c0c7\_MG\_R\_ML\_ANON & pos/22678953\_b9a4da5f2dae63a9\_MG\_R\_CC\_ANON \\pos/22679008\_b9a4da5f2dae63a9\_MG\_R\_ML\_ANON & pos/24065707\_5291e1aee2bbf5df\_MG\_R\_ML\_ANON \\pos/24065761\_5291e1aee2bbf5df\_MG\_R\_CC\_ANON & pos/30011674\_6968748e66837bc7\_MG\_L\_CC\_ANON \\pos/30011727\_6968748e66837bc7\_MG\_L\_ML\_ANON & pos/30011824\_4f20c1285d8f0b1f\_MG\_L\_CC\_ANON \\pos/30318067\_4f20c1285d8f0b1f\_MG\_L\_ML\_ANON & pos/50998059\_66adfbb4f19c76d2\_MG\_L\_ML\_ANON \\pos/50998113\_66adfbb4f19c76d2\_MG\_L\_CC\_ANON & neg/20588680\_036aff49b8ac84f0\_MG\_L\_ML\_ANON $^\ast$\\
         \hline 
    \end{tabular}
    \caption{Image names of the test examples from the INbreast data set, as used in Shen et al.~\cite{shen2019deep}. $^\ast$This image was not included in the original test set. It was added to prevent any data leakage since another image from the same exam/patient is included in the training set.}
    \label{tab:supplement_inbreast}
\end{center}
\end{table}

\end{document}